\documentclass[conference]{IEEEtran}
\usepackage{svg}
\IEEEoverridecommandlockouts
\usepackage{cite}
\usepackage{amsmath,amssymb,amsfonts}
\usepackage{algorithmic}
\usepackage{graphicx}
\usepackage{textcomp}
\usepackage{xcolor}
\usepackage{tikz}
\usepackage{multirow}
\usepackage{afterpage}
\usepackage{placeins}
\usepackage{url}
\usepackage{cleveref}
\usepackage{float}
\def\BibTeX{{\rm B\kern-.05em{\sc i\kern-.025em b}\kern-.08em
    T\kern-.1667em\lower.7ex\hbox{E}\kern-.125emX}}
\usepackage[most]{tcolorbox}

\begin{document}

\title{Enhancing Player Enjoyment with a Two-Tier DRL and LLM-Based Agent System for Fighting Games
\\
}

\author{\IEEEauthorblockN{1\textsuperscript{st} Shouren Wang}
\IEEEauthorblockA{\textit{Game Innovation Lab, NYU} \\
\textit{Case Western Reserve University}\\
Cleveland, USA \\
sw5004@nyu.edu}
\and
\IEEEauthorblockN{2\textsuperscript{nd} Zehua Jiang}
\IEEEauthorblockA{\textit{Game Innovation Lab} \\
\textit{New York University}\\
New York, USA \\
zehua.jiang@nyu.edu}
\and
\IEEEauthorblockN{3\textsuperscript{rd} Fernando Sliva}
\IEEEauthorblockA{\textit{Electronic Arts} \\
New York, USA \\
fms2005@gmail.com}
\and
\and
\IEEEauthorblockN{4\textsuperscript{th} Sam Earle
}
\IEEEauthorblockA{\textit{Game Innovation Lab} \\
\textit{New York University}\\
New York, USA  \\
sam.earle@nyu.edu}
\and
\IEEEauthorblockN{5\textsuperscript{th} Julian Togelius}
\IEEEauthorblockA{\textit{Game Innovation Lab} \\
\textit{New York University}\\
New York, USA  \\
julian@togelius.com}
}

\maketitle

\begin{abstract}
Deep reinforcement learning (DRL) has effectively enhanced gameplay experiences and game design across various game genres. However, few studies on fighting game agents have focused explicitly on enhancing player enjoyment, a critical factor for both developers and players. To address this gap and establish a practical baseline for designing enjoyability-focused agents, we propose a two-tier agent (TTA) system and conducted experiments in the classic fighting game Street Fighter II. The first tier of TTA employs a task-oriented network architecture, modularized reward functions, and hybrid training to produce diverse and skilled DRL agents. In the second tier of TTA, a Large Language Model Hyper-Agent, leveraging players' playing data and feedback, dynamically selects suitable DRL opponents. In addition, we investigate and model several key factors that affect the enjoyability of the opponent. The experiments demonstrate improvements from 64. 36\% to 156. 36\% in the execution of advanced skills over baseline methods. The trained agents also exhibit distinct game-playing styles. Additionally, we conducted a small-scale user study, and the overall enjoyment in the player's feedback validates the effectiveness of our TTA system.
\end{abstract}


\section{\textbf{Introduction}}

Fighting games are a competitive game genre that has been popular since the arcade era. Their defining characteristic is the requirement for players to execute actions or a series of complex maneuvers with speed and precision. The origins of fighting games can be traced back to arcade halls, with classic titles such as Street Fighter II (SF2) and The King of Fighters (KOF) '97. Over time, the genre has evolved to include modern titles like Street Fighter 6, the Mortal Kombat series, and the Super Smash Bros. series. Due to their real-time nature and competitive mechanics, fighting games are inherently well-suited for research in game AI systems, particularly in the field of reinforcement learning (RL). Among all fighting games, the Street Fighter series stands out as the most renowned and widely recognized pioneer of the genre, making it the ideal testbed for AI research in fighting games.

In recent years, several studies have emerged on fighting games, most of them based on Deep Reinforcement Learning (DRL). The DRL methods have proved to be very effective in commercial fighting games, Naruto Mobile, their DRL agents can outperform most human players and have been applied to the online game mode and effectively attracted new players\cite{naruto_1_liu2023naruto, naruto_2_zhang2024advancing}. However, Naruto Mobile can hardly be considered a strict fighting game, its core mechanics are closer to action games.  In contrast to traditional fighting games like SF2 and KOF, Naruto Mobile features simplified controls (skills with a single button press) and unbalanced characters, making it less challenging for game-playing agents. This limits the applicability of their approach to more complex fighting games. As a result, the methods developed for Naruto Mobile are unlikely to be directly applicable to traditional fighting games. 


On the FightingICE platform, extensive research has been done on fighting game agents, including agents that take advantage of Monte Carlo Tree Search (MCTS) to achieve strong performance\cite{FightingICE_kim2020mastering}. However, compared to end-to-end methods that directly output action probabilities, MCTS incurs significantly higher computational costs~\cite{mcts_lodel2022look}, making it unsuitable for real-time fighting games, especially AI in commercial games where efficiency and inference speed are crucial. Although, as an open-source platform, FightingICE provides full access to environment information via its API, compared to commercial fighting games, it falls short in terms of the number of playable characters, game balance, and overall game mechanism, which are crucial to fighting games.


As a pioneer in fighting games, SF2 is a highly suitable testbed for AI research in fighting games. Several studies have been conducted on Street Fighter game-playing agents. The double Deep-Q Network (DDQN) method has been applied to Street Fighter V and has achieved a desirable win rate\cite{sf_liang2022study}. Furthermore, DRL approaches using self-play training and proxy policy optimization (PPO) have been used in SF2 within the OpenAI Gym Retro(Retro) environment\cite{sf_go2023phase}. However, these studies focus primarily on the performance of DRL agents—namely, their win rate—rather than whether the play styles of these agents enhance the player's enjoyment. Although performance is usually the most critical metric in other DRL application fields, such as robotics, in video games, the player's enjoyment during battles is ultimately more important than the AI's raw strength. An agent optimized only to win, especially if optimized against a specific opponent or specific scenario, may learn a highly specific policy that, while effective, is uninteresting to play against.

To enhance player enjoyment, we propose the two-tier agent (TTA) system, designed to provide diverse game-playing experiences in the Champion Edition of SF2. Our TTA system achieves this by introducing opponents with distinct game-playing styles, capable of executing advanced skills, for example, special moves, and dynamically selecting opponents based on players' behavior and feedback. The first tier of TTA contains an agent archive, which contains different types of DRL game-playing agent that exhibit distinct game-playing game-playing styles. To obtain a well-curated agent archive, we propose a carefully designed model architecture for the SF2 game, a modularized reward function with multiple reward terms and a hybrid self-play training method. These approaches enable agents not only to learn advanced techniques, such as special moves, but also to develop diverse fighting styles, including defensive, special move-focused, and newbie-type agents. The second tier of TTA contains a game manager (GM) to collect the playing data and run the game, and a Large Language Models hyper-agent (LLMHA) that selects opponents from the DRL agent archive based on player's playing data and feedback. 



In summary, our contributions in this paper are as follows:
\begin{itemize}
    \item We present a DRL and LLMs based agents system, TTA, to improve player's enjoyment.
    \item We proposed a well-designed DRL model architecture, a modularized reward function, and a training method to generate DRL agents with different game styles and can use advanced skills.
    \item We applied well-designed prompt for LLMHA to dynamically select opponents for players to improve players' enjoyment.
\end{itemize}

Due to page limits, additional experimental details and supplementary results are provided in our GitHub repository\footnote{\url{https://github.com/WangShourenWSR/fighting-game-two-tier-agent-system}}.

\section{\textbf{Preliminaries}}
In this section, we introduce the game environment, formulation of the RL problem, and methodological background.
\subsection{Game Environment}
\subsubsection{Street Fighter II: Champion Edition}\label{game_sf2}
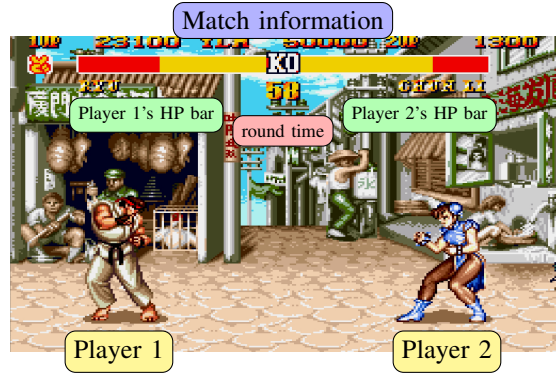
\begin{figure}
    \centering
    \begin{tikzpicture}
        \node[anchor=south west,inner sep=0] (img) at (0,0) {\includegraphics[width=0.4\textwidth]{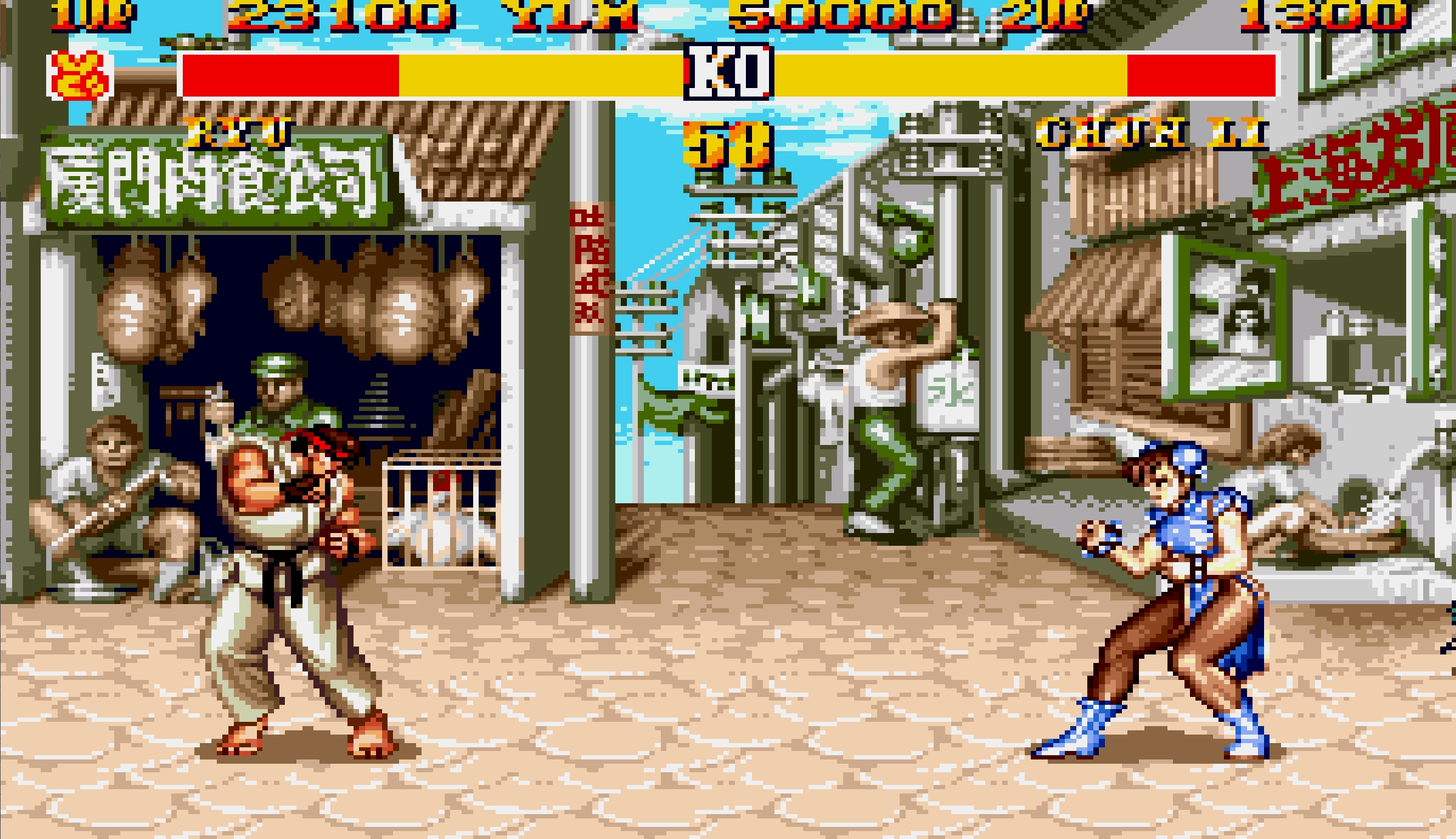}};
        \begin{scope}[x={(img.south east)},y={(img.north west)}]
            \node[draw,fill=yellow!50,rounded corners] at (0.2,0.0) {Player 1};
            \node[draw,fill=yellow!50,rounded corners] at (0.8,0.0) {Player 2};
            \node[draw,fill=green!30,rounded corners] at (0.25,0.75) {\scriptsize  Player 1's HP bar};
            \node[draw,fill=green!30,rounded corners] at (0.75,0.75) {\scriptsize  Player 2's HP bar};
            \node[draw,fill=red!30,rounded corners] at (0.50,0.70) {\scriptsize round time};
            \node[draw,fill=blue!30,rounded corners] at (0.50,1.05) { Match information};
        \end{scope}
    \end{tikzpicture}
    \caption{Street Fighter II Game Interface: Champion Edition. The labeled Player 1 (left) and Player 2 (right) indicate the characters controlled by each player; health point (HP) bars display each player's remaining health, with yellow representing the current HP and red indicating lost HP; round time in the center determining the remaining duration of the round; match information at the top provides the match information, for example, score.}
    \label{fig:game_pixel}
\end{figure}

Street Fighter II: Champion Edition has 12 balanced and diverse characters, each with unique abilities, including their distinct special moves. The battles follow a one-on-one format and are played in a 2D plane as Fig.~\ref{fig:game_pixel} shows. Players can use ten buttons to control jumping, crouching, moving left and right where players move left, right, jump, crouch and light, medium, and heavy punches and kicks. The round ends when one of the character's health points (HP) reaches zero or time runs out. 

Worth noting is that SF2's special move system transforms a seemingly simple 10-button input scheme into a deep and intricate skill execution mechanic. For example, performing Ryu’s Shoryuken, an anti-air special move with a few frames of invincibility, requires precisely inputting '→' + '↓' + '→' + 'punch' within a tight frame window. This intricate input system makes RL tasks for SF2 much more challenging than common action games, requiring DRL agents' to perform temporally extended sequences rather than simply instantaneous actions
(more details are discussed in Section~\ref{method: network_architecture}).


\subsubsection{OpenAI Gym Retro}
Extended from the standard OpenAI Gym, Retro is an open source platform designed to facilitate reinforcement learning (RL) research using classic video games. Retro granting access to thousands of games, including SF2. Researchers can explore memory addresses to extract game data, adapting it to the specific requirements of their RL tasks. 


\subsection{Task Formulation}
\label{task_formulation}

According to the reinforcement learning framework of Sutton and Barto \cite{rl_sutton_barto}, the SF2 task can be formulated as a Markov Decision Process (MDP). The MDP of Street Fighter II can be described as a five-element tuple $\{S, A^{a}, A^{o}, R, T \}$, where $S$ is the state space, $A^{a}$ is the action space of the agent, $A^{o}$ is the action space of the opponent, $R$ is the reward function, and $T$ is the environment dynamics of the game. In each time step $t$, the agent's policy $\pi^{a}$ and the opponent's policy $\pi^{o}$ will sample their actions $a_t^a$ and $a_t^o$ according to the state $s_t$.
\begin{equation}
    a_t^a \sim \pi^a(a_t^a | s_t) , a_t^o \sim \pi^o(a_t^o | s_t)
\end{equation}
The state transition is executed by the emulator integrated in Retro, which is deterministic, thus, the game environment dynamics $T$ is deterministic rather than stochastic, which means state in time step $t+1$ is determined if the state and actions in time step $t$ is known.
\begin{equation}
    s_{t+1} = T(s_t, a_t^a, a_t^o)
\end{equation}
There are two types of tasks in our proposed training method, PvE and self-play. In PvE tasks, the opponent policy $\pi^o$ is a deterministic built-in AI designed by CAPCOM; In self-play tasks, the opponent policy $\pi^o$ is a historical DRL agent model that generates a probability distribution over actions. If $\pi^o$ is fixed, the opponent can be reclassified as part of the environment: for PvE tasks, the new environment dynamic $T^{PvE}$ is still deterministic, but for self-play tasks, the environment dynamics becomes stochastic and is represented by a probability distribution $P$.
\begin{equation}
    s_{t+1}=
    \begin{cases} 
        T^{PvE}(s_t, a_t^a), & $PvE tasks$ \\
        P(s_t, a_t^a), & $self-play tasks$
    \end{cases}
    \label{equation_env_dynamics}
\end{equation}
Thus, SF2's MDP can be simplified as $\{S, A^{a}, R, T^{PvE}\}$ for PvE tasks, and $\{S, A^{a}, R, P\}$ for self-play tasks. 

We designed multiple reward functions for SF2 tasks to guide agents to learn diverse game-playing styles. The agent's goal is to optimize its policy $\pi^a$  to the optimal $\pi^{a*}$ to maximize the expected cumulated reward:
\begin{equation}
    J(\pi^a) =  \mathbb{E} \Big[ \sum_{t=0}^{T} \gamma^t R_t \Big]
\end{equation}
where $\gamma$ is the discount factor and $R_t$ is the reward generated by our proposed modularized reward function with keyword arguments $\lambda$ as input(see Section\ref{method: reward_function}).
\begin{equation}
    R_t  =  r(s_t, a_t^a, \lambda)
    \label{equation_reward}
\end{equation}

\subsection{Methodological Background}
\subsubsection{Deep Reinforcement Learning}
Traditional tabular RL methods rely on explicit storage and lookup of state-action values to determine an agent's policy, which are effective in small, discrete state spaces, but become infeasible in complex RL tasks\cite{rl_sutton_barto}.

To address this limitation, DRL replaces explicit tables with deep neural networks which have the capability to approximate highly complex nonlinear functions. In our task, the agent policy $\pi^a$ is modeled by a deep neural network with parameter $\theta$. We applied Proximal Policy Optimization (PPO)\cite{ppo_schulman2017proximal} as the policy optimization method for the policy network. We used stable-baselines3\cite{stable-baselines3} as the implementation of PPO.

\subsubsection{Large Language Models}
Large Language Models (LLMs) are deep neural networks pre-trained through self-supervised learning on large-scale text corpora and further refined through fine-tuning. Built on multilayer transformer\cite{attn_is_all_you_need_vaswani2017attention}, such as GPT-3 with 175 billion parameters\cite{gpt3_brown2020language}, LLMs contain billions of parameters, far exceeding traditional models like Convolutional Neural Network (CNN). This scale parameter enables LLMs to excel in various tasks of natural language and reasoning.

One key advantage of LLMs is their ability to perform In-Context Learning (ICL)\cite{gpt3_brown2020language}, where models generalize new tasks by conditioning on prompts without explicit weight updates. Among the latest LLM optimized for reasoning, DeepSeek-R1 has demonstrated state-of-the-art reasoning capabilities, in particular integrates advanced CoT and RL techniques\cite{r1_guo2025deepseek}, making it highly suitable as our hyper-agent.

\subsubsection{Hyper-agent}
A hyper-agent is a meta-level decision system that selects the most appropriate sub-agent for a given task, rather than acting directly. It has been applied in algorithm selection and adaptive decision-making in various domains, including general video game playing\cite{hyper_agent_mendes2016hyper} and ad hoc cooperation\cite{hyper_agent_canaan2022generating}.

Reasoning-optimized LLMs excel in decision-making, making them ideal Hyper-Agents. Through ICL prompting, sub-agent selection rules can be defined directly in natural language. Additionally, LLMs' ability to directly understand natural language allows them to take player feedback as input without extra processing, giving them a unique advantage over traditional Hyper-Agents in handling player feedback. In our approach, DeepSeek R1 serves as the hyper-agent, leveraging ICL prompts combined with players' playing data and feedback to dynamically select suitable opponents, enhancing player's enjoyment (see Section~\ref{method: LLMHA} and Section~\ref{Exp: user_study}).

\subsection{Hypothesis on Player Enjoyment in Competitive Matches}
\label{prelimin: Player Enjoyment}

Player enjoyment is a subjective experience, but certain objective patterns and commonalities can be identified. The literature on what makes games enjoyable, in general and for particular players, is relatively well-developed and diverse~\cite{lazzaro2009we,pedersen2009modeling,koster2004theory,malone1981makes}. Previous research in the game industry has examined how different types of opponents influence the enjoyment of players in competitive matches\cite{enjoyment_Activision2024, enjoyment_gutwin2016peak}. Based on these industry studies and empirical experience as players, we hypothesize that the following factors significantly impact player enjoyment in a series of matches:

\begin{enumerate}
    \item \textbf{Opponent Skill Level}: Generally, players have greater enjoyment from balanced matches with opponents of similar skill levels. The satisfaction of winning a close and intense match against an evenly matched opponent is significantly higher than easily defeating a weaker opponent or being overwhelmingly defeated by a much stronger one.
    
    \item \textbf{Opponent Use of Advanced Techniques}: Players tend to enjoy matches more when their opponents are able to utilize advanced strategies and techniques. If an opponent lacks such skills, even a victory may feel unfulfilling. A worse scenario arises when an opponent easily defeats the player without advanced techniques, leading to a sense of humiliation, often called ``smurfing."
  
    \item \textbf{Opponent Diversity}: Players are more likely to enjoy matches when they face opponents with diverse game-playing styles. Repeated encounters with opponents employing similar tactics can lead to a monotonous and less engaging experience.
    
    \item \textbf{Matchmaking System}: A well-designed matchmaking system that selects appropriate opponents based on player data, ensuring balanced skill levels, varied game-playing styles, and the use of advanced techniques, can significantly enhance player enjoyment.
\end{enumerate}

\section{\textbf{Method}}
\begin{figure*}
    \centering
    \includegraphics[width=1.0\linewidth]{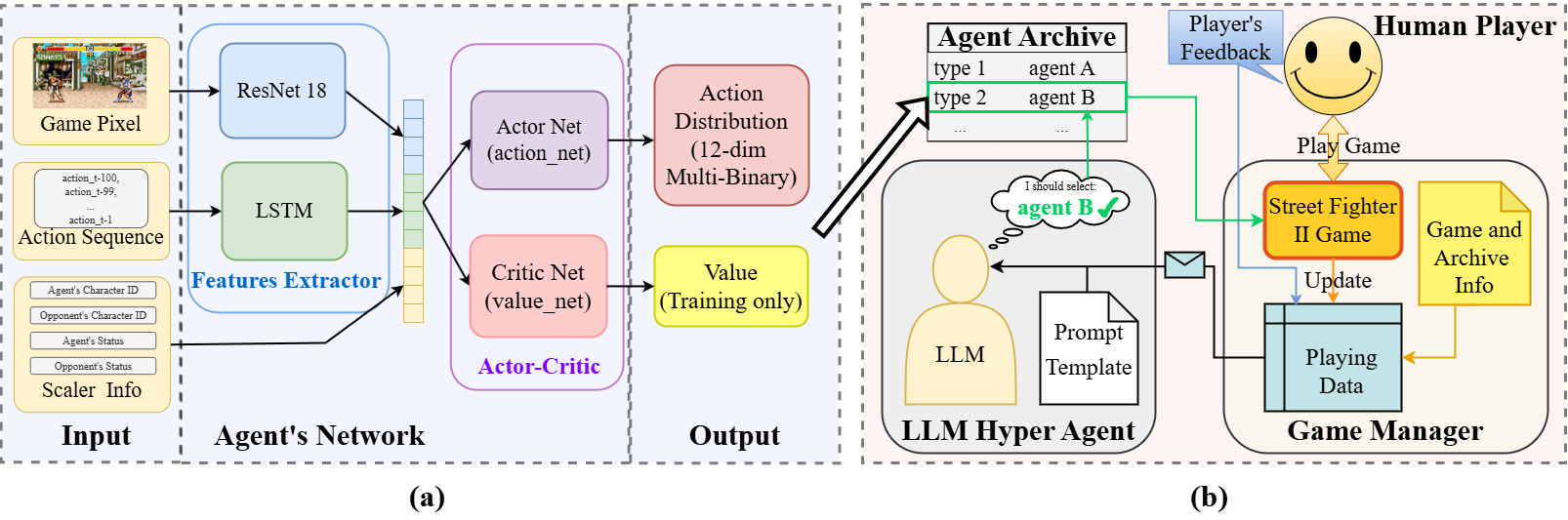}
    \caption{Overview of the TTA system: (a) Network architecture of the DRL agent; (b) LLM Hyper-Agent opponent selection process. 
    (a) The network processes three types of inputs: game pixels, scalar information, and an action history spanning the past \( 100\) steps. The extracted features are fed into an actor-critic network, where the actor net (MLP) produces a 12-dimensional multi-binary action probability distribution, and the critic net estimates the value function (used only during training). The feature extractor consists of a CNN module (ResNet18) for visual feature extraction and an RNN module (LSTM) for learning sequential dependencies, particularly for executing special moves. Scalar information, including character ID and game states, is concatenated with the extracted features before being processed by the actor-critic network. 
    (b) LLMHA selection pipeline, which dynamically selects DRL agent opponents for the player based on their match feedback and playing history. The GM maintains a record of the player's playing data (e.g., win rate, previous opponents) and, after each match, prompts the player for feedback. This feedback is then integrated into the playing data and passed to the LLMHA. The LLMHA embeds the playing data into a prompt template and uses it to infer the most suitable opponent, ensuring an adaptive and personalized experience.}
    \label{fig:TTA_system}
\end{figure*}
In this section, we introduce the design and methods of our TTA system, which consists of two tiers: the DRL game-playing agents tier and the Hyper-Agent tier. The overall architecture and system workflow are illustrated in Fig.~\ref{fig:TTA_system}.  
\begin{itemize}
    \item DRL game-playing agents tier: Network architectures for DRL agents, the design of the modularized reward function for diverse types of agents, and the hybrid training that combined both self-play tasks and PvE tasks.
    \item Hyper-agent tier: The agent archive that stores the DRL game-playing agents, the game manager (GM), and the prompt design for LLMHA.
\end{itemize}

\subsection{Network Architecture}\label{method: network_architecture}
Fig.~\ref{fig:TTA_system}~(a) illustrates the structure of our network, which processes three types of input: images (game pixels), scalar data (in-game numerical states) and sequential data (action sequences from previous \( n \) steps). The model outputs a 12-dimensional multi-binary action probability distribution corresponding to 12 game buttons and a scalar value which is used only during training to compute the PPO loss.

The network of DRL agents consists of two main components: the features extractor and the actor-critic network. The actor-critic network includes an action network (actor) and a value network (critic), both implemented as multilayer perceptrons (MLP). The actor outputs the probability distribution of the action and the critic estimates the value function.

The feature extractor is made up of two modules: a CNN extractor and a recurrent neural network (RNN) extractor. The CNN extractor processes game pixels, extracting visual features and producing a feature vector. Here, we adopt ResNet18\cite{resnet_he2016deep} without modification, as the CNN extractor. Since executing special moves requires precise execution of a series of actions (see Section \ref{game_sf2}), our model incorporates an RNN to extract temporal features from the agent's action history in the past \( n \) steps, producing a feature vector; in our experiments, we set \( n \) $= 100$. Here, we adopt the Long-Short-Term Memory (LSTM) architecture\cite{lstm_hochreiter1997long} as our RNN extractor. Additionally, we incorporate scalar in-game state information, including the current states of the agent and opponent (e.g., attacking, being hit, stunned), as well as their character IDs. These scalar features are concatenated with the feature vectors from the CNN and RNN extractors. The concatenated vector is then fed into the actor-critic network. Through forward propagation, the DRL agent network outputs the action probability distribution and value.

\subsection{Modularized Reward Function}
\label{method: reward_function}
\subsubsection{Modularized Reward Function}

As assumed in the preliminaries (see Section~\ref{prelimin: Player Enjoyment}), most players prefer opponents with diverse game-playing styles and can execute advanced techniques. Therefore, the agent archive should consist of various DRL agents, each exhibiting a particularly distinct game-playing style, with most agents capable of performing advanced skills. To enable the learning of these game-playing and techniques, we design a modularized reward function.

Our modularized reward function comprises multiple components: Raw reward, the fundamental reward based on changes in HP and the manner in which damage is dealt (e.g., damage dealt by special moves); Special move reward, where agents can receive additional rewards for performing special moves; Projectile reward, which grants rewards for using projectile attacks; Distance reward, given for maintaining a specific distance from the opponent; In-air reward, where staying in the air contributes to the total reward; Time reward, where the duration of a match influences the reward received; and Cost, where certain actions incur penalties when executed. Each component is computed individually, and the final reward returned to the agent (as defined in Equation\eqref{equation_reward}) is obtained by adding these components.


The modularized reward function takes two key parameters:

\begin{itemize}
    \item \textbf{Info dictionary from the environment}: This dictionary is analyzed to determine whether the agent has performed specific actions or entered certain states, such as executing a special move.
    \item \textbf{Customizable reward terms dictionary}: This dictionary contains coefficients for various reward components. For example, if the coefficient for the special move reward is positive, the agent receives a reward when executing a special move.
\end{itemize}

For each desired DRL agent type, we define a specific reward configuration using a specialized reward terms dictionary. This modularized reward function enables the training of various agent types, including \textit{projectile-type}, \textit{special-move-type}, \textit{defensive-type}, \textit{air-type}, \textit{coward-type}, \textit{newbie-type}, and \textit{key-spamming-type}. Reward terms settings for each type of DRL agents can be found in Appendix~\ref{appd_reward_term_settings}. These trained agents are stored in the agent archive of the TTA system. 

\subsection{Hybrid Training}
The Arcade Mode in SF2 includes built-in AI opponents designed by CAPCOM. These rule-based AIs are capable of executing advanced techniques, making them suitable opponents for training DRL agents. However, using only built-in AI as opponents, a setting which we refer to as PvE tasks, to train the agent, has a critical limitation: Because the game emulator is deterministic (see Section~\ref{task_formulation} and Equation~\eqref{equation_env_dynamics}), after multiple training iterations, the DRL agent tends to overfit to a fixed pattern, becoming a quasi-open-loop controller that operates almost without feedback and executes actions based solely on fixed inputs. This has been proved in an SF2 project with PvE tasks only\cite{linyi_streetfighterai}. Such an agent, functionally equivalent to a macro, is completely unusable.

Most existing research on fighting game agents relies on self-play for training. However, in complex RL tasks like Street Fighter, we observed that due to the reward function design and self-play dynamics, the agent's policy may converge to a locally optimal. In our experiments, this manifested itself as persistent crouch kick or jump kick spamming, preventing the agent from learning advanced techniques.

To address these issues, we propose Hybrid Training, which combines the built-in AI of the Arcade Mode with self-play in a balanced ratio to improve the agent's performance (see Section~\ref{experiment_setup} for more details).

\subsection{Hyper-Agent Tier}
\label{method: LLMHA}
As the second tier of the TTA system, the Hyper Agent tier consists of three key components:

\begin{itemize}
    \item \textbf{Agent Archive}: A collection of DRL agents categorized by different game-playing styles.
    \item \textbf{Game Manager}: Maintains the agent archive, records player data, ask players for feedback, and interacts with the LLMHA to select the next opponent for the player. It serves as the intermediary between the agent archive, the player, the game, and the LLMHA.
    \item \textbf{LLM Hyper Agent}: Integrates the information from the GM with a prompt template and uses a LLM to infer the next opponent. The LLMHA then extracts the selection from the LLM's output and returns it to the GM.
\end{itemize}
The key design considerations for this system include the playing data maintained by the GM and the prompt template used by the LLMHA.

The GM records the player data including: player's character, win rate, total wins/losses, current win/loss streak, the opponent in the last match, and player's behavior data (e.g. special move usage). Additionally, after each round, the GM asks the player for optional feedback. If feedback is provided, the GM stores the feedback along with information about the previous opponent. Before each match, the GM sends the player's data to the LLMHA to determine the next opponent.

The LLMHA prompt template consists of six components: (1)~base template, (2)~selection principles, (3)~output format requirements, (4)~ICL examples, (5)~archive information, (6)~playing data. Components (1), (2), (3) and (4) are predefined, and (5) is provided by the GM during the initialization of LLMHA, and (6) is provided by the GM dynamically. LLMHA combines these components into a unified prompt as input to the LLM. The LLM's output is then processed to extract a candidate opponent and return it to the GM. If the LLM does not generate a valid selection, the LLMHA re-queries its LLM for a new inference. The complete structure and example of the prompt template are provided in Appendix~\ref{appd_prompt_templates}, and a LLMHA's output example is provided in Appendix~\ref{appd_llmha_output}

\section{Experiments}
In this section, we present the experimental setup and results of the TTA system. First, we detail the configuration of hyperparameters for DRL agents, including network architectures, reward functions, hybrid training pipelines, and  prompt design for the LLMHA. Next, we introduce the evaluation methods and a comprehensive analysis of the experimental results for DRL agents, LLMHA, and overall player enjoyment. 
\subsection{Experiment Setup}\label{experiment_setup}
\subsubsection{DRL Agent Network Hyper Paramters}
\begin{table}
    \caption{Network parameter}
    \begin{center}
        \begin{tabular}{c c|c c|c c}
            \hline
            \multicolumn{2}{c|}{\textbf{LSTM (RNN)}} & \multicolumn{2}{|c|}{\textbf{Actor Network}} & \multicolumn{2}{|c}{\textbf{value Network}}\\
            \hline 
            \textbf{Parameters} & \textbf{Values} & \textbf{Parameters} & \textbf{Values} & \textbf{Parameters} & \textbf{Values}\\
            \cline{1-6}
            {Input dim} & {12} & {Layer1 dim} & {512}& {Layer1 dim} & {512}\\
            {Hidden dim} & {128} & {Layer2 dim} & {256} &  {Layer2 dim} & {256} \\
            {Num layers} & {2} & {Layer3 dim} & {128} & {Layer3 dim} & {128}\\
            {Batch first} & {True}  & {Layer4 dim} & {128} & {Layer4 dim} & {128}\\
            {Dropout} & {0.1} & {Output dim} & {12} & {Output dim} & {1}\\
            {} & {} &  {Activation} & {Tanh}&{Activation} & {Tanh} \\
            \hline
        \end{tabular}
        \begin{flushleft}
        \centering
            \footnotesize The CNN module is default ResNet18 and not listed here.
        \end{flushleft}
        \label{table_network_hyperparameter}
    \end{center}
\end{table}

        \begin{table*}[htbp]
            \caption{Reward Terms}
            \begin{center}
                \begin{tabular}{c|c}
                    \hline 
                    \textbf{Parameters} & \textbf{Description} \\
                    \hline
                    {Reward scale}  & {Scale the reward by multiplying it by this term} \\
                    {Raw reward coefficient} & {Scale the HP-based reward by multiplying it by this term} \\
                    {Special move bonus}  & {Scale the HP-based reward caused by special moves by multiplying it by this term}  \\
                    {Projectile bonus}  & {Scale the HP-based reward caused by projectiles by multiplying it by this term}  \\
                    {Distance bonus}  & {Scale the HP-based reward according to distance between agent and opponent by multiplying it by this term} \\
                    {Special move reward}  & {Reward for triggering a special move} \\
                    {Projectile reward} & {Reward for triggering a projectile} \\
                    {Distance reward}  & {Positive/Negative value for reward keeping a long/close distance from the opponent}  \\
                    {In air reward}  & {Reward for being in air} \\
                    {Time reward}  & {Positive/Negative value for reward take more/less time for each round}  \\
                    {Cost coefficient}  & {Scale the cost (penalty) by multiplying it by this term}  \\
                    {special move cost} & {Cost for triggering a special move} \\
                    {Regular attack cost}  & {Cost for triggering a regular attack} \\
                    {Jump cost} & {Cost for triggering a jump} \\
                    {Vulnerable frame cost}  & {Cost for being in a vulnerable frame, i.e. frames that agent cannot control the character} \\
                    \hline
                \end{tabular}
            \begin{flushleft}
            \centering
                \footnotesize The description of modules in the modularized reward function.
            \end{flushleft}
            \label{table_reward_terms_default}
            \end{center}
    \end{table*}
We employ unmodified ResNet-18\cite{resnet_he2016deep} as the network architecture of the CNN feature extractor. TABLE~\ref{table_network_hyperparameter} lists the hyperparameters for the other parts of the DRL agents' network. For the PPO algorithm, we set the value function coefficient to 1, and the entropy coefficient to 0.01. We also adopt a learning rate scheduler in which the learning rate decreases linearly from $2.5 \times 10^{-4}$ to $2.5 \times 10^{-6}$. For hybrid training, we trained the DRL agents with 30\% PvE tasks and 70\% self-play tasks where agents compete against all their historical versions. Worth mentioning, we set a 50\% character flip rate to swap the agent's and opponent's characters in each round to further enhance the robustness by avoiding the effect of initial left and right positions of player 1 and player 2 in the game. To speed up training, we set up 12 parallel environments to collect RL rollout data and train the DRL agent.
TABLE~\ref{table_reward_terms_default} lists the reward terms and their descriptions. The reward term settings for each type of reward are provided in Appendix~\ref{appd_reward_term_settings} 

\subsubsection{Hyper Agent Setup}
Considering cost efficiency, we selected DeepSeek-R1 from the open-weight models available. As a reasoning model, DeepSeek-R1 exhibits strong reasoning capabilities, making it well suited for analyzing player behavior and feedback to support opponent selection within the Hyper Agent framework.

For deployment, we opt for the DeepSeek-R1-Distill-Llama-8B variant and deploy it locally. This lightweight model outperforms DeepSeek-R1-Distill-Qwen-1.5B in performance while maintaining computational efficiency. Under appropriate configurations, it can run inference smoothly on desktop- and even laptop-grade local machines.

Our local machine is equipped with an Nvidia RTX 3080 laptop GPU with 16GB of VRAM. By applying 8-bit quantization, DeepSeek-R1 can operate efficiently on this hardware, enabling real-time inference within our experimental setup.
\subsection{Evaluation and Results}
To evaluate the impact of TTA on player enjoyment, we establish metrics for assessing both the agent's gameplay diversity and the effectiveness of LLMHA in opponent selection.

\subsubsection{Mastery of Advanced Techniques}
\begin{figure}
    \centering
    \includegraphics[width=1.0\linewidth]{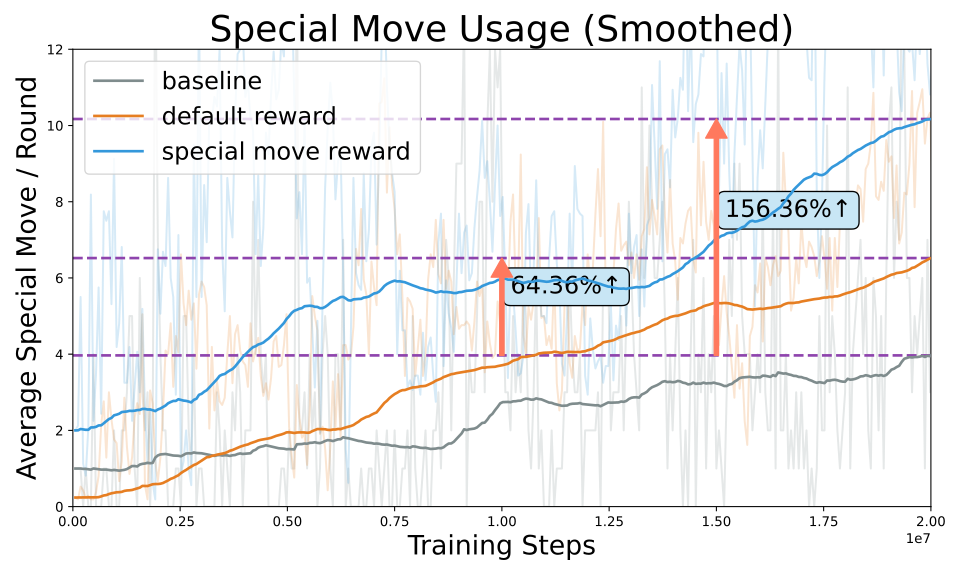}
    \caption{The figure shows the average number of special moves performed per round by DRL agents trained with three approaches. The ``baseline'' method uses a CNN+MLP architecture with HP-based rewards\cite{sf_go2023phase} The ``default reward'' method adopts our proposed DRL network while keeping HP-based rewards. The ``special move reward'' method further incorporates a reward terms tailored for special moves (see Section~\ref{method: reward_function}). Compared to the Baseline, the ``default reward'' method improves special move usage by 64. 36\%; the ``Special Move Reward" method achieves a 156. 36\% increase, demonstrating the effectiveness of our network architecture and reward design.}

    \label{fig:result_special_move}
\end{figure}
\begin{figure}
    \centering
    \includegraphics[width=1.0\linewidth]{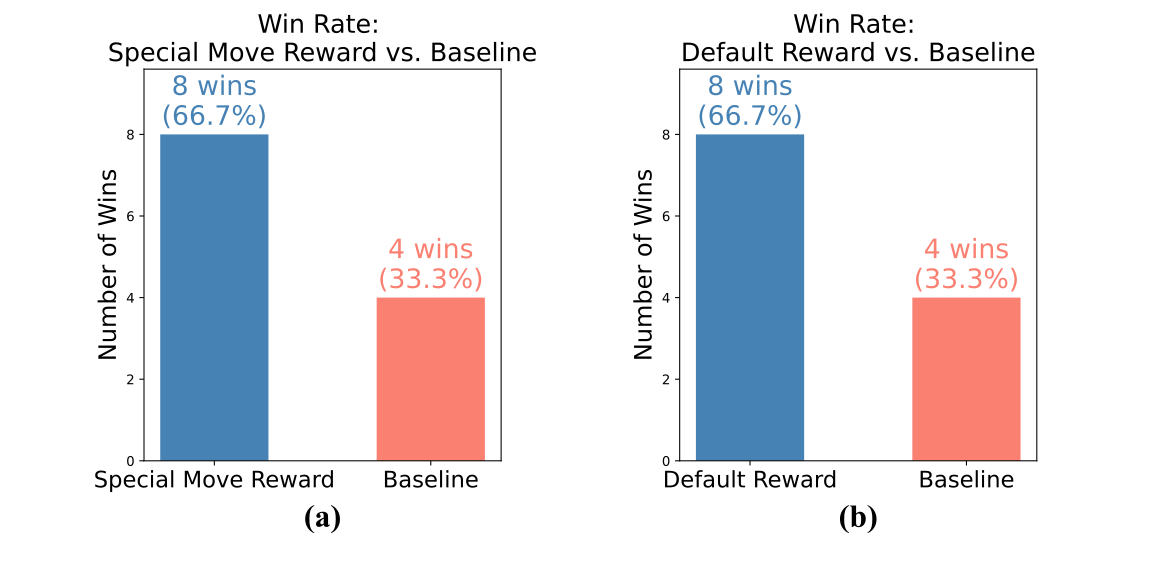}
    \caption{Win rate comparison between our DRL agent and the baseline model~\cite{sf_go2023phase}. The figure presents the results of 12 matches. Regardless of the reward function used, our model consistently outperformed the baseline with a 66.7\% win rate. This demonstrates that our network architecture significantly improves the DRL agent's performance in competitive matches.}
    \label{fig:result_win_rate}
\end{figure}
Fig.~\ref{fig:result_special_move} illustrates the ability of DRL agents to utilize advanced techniques. We compare DRL agents trained with our proposed model architecture and modularized reward functions with those trained with the same architecture but using a default reward function, as well as agents based on CNN + MLP at baseline\cite{sf_go2023phase}. The results show that, compared to the baseline method, our DRL agents trained using special-move- preferred and default HP-based reward settings improve the use of special move by 64. 36\% and 156. 36\%, respectively. This result validates the effectiveness of both the model architecture and the design of modularized reward.

Furthermore, mastering advanced techniques is only a necessary condition for achieving high combat performance (i.e., win rate). To further validate the effectiveness of our model in actual matches, we conducted direct battles between our trained agents and the baseline model. Fig.~\ref{fig:result_win_rate}~(a) presents the results of agents trained with the special move reward competing against the baseline model, while Fig.~\ref{fig:result_win_rate}~(b) shows the results for agents trained with the default reward. Both tests contain 12 matches, where both agents in the test use the same character in each match, covering all 12 available characters. The results of both tests show that our model outperforms the baseline with a substantial 66.7\% win rate advantage, demonstrating its effectiveness and confirming that our model structure excels in competitive performance.
\subsubsection{Distinct game-playing styles}
\begin{figure}
    \centering
    \includegraphics[width=1.0\linewidth]{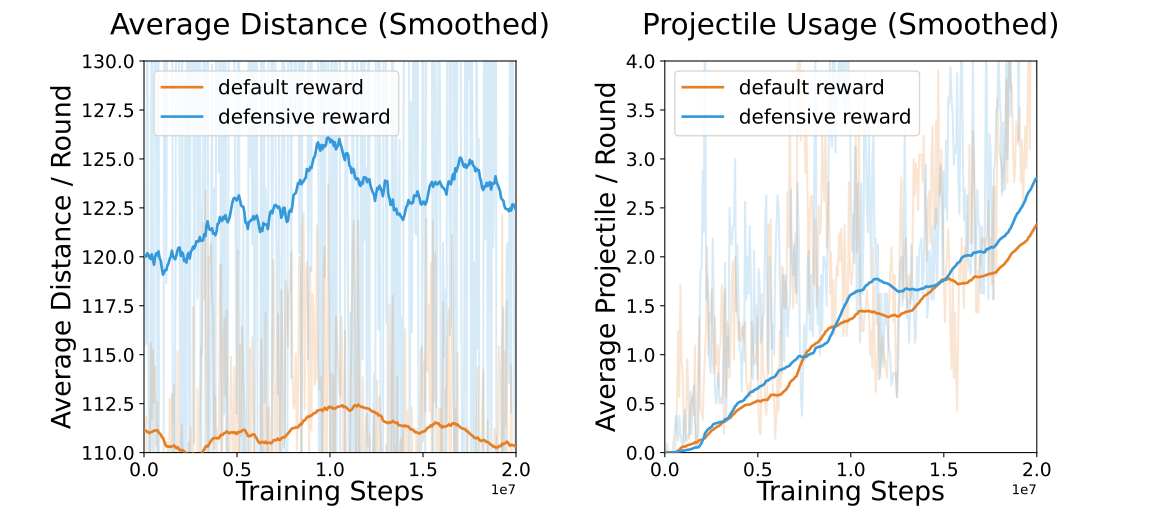}
    \caption{Comparison of key playing behaviors between DRL agents trained with the default reward and the defensive reward. The figure presents two behavioral metrics: average distance from the opponent and projectile usage rate. The defensive reward agent maintains a noticeably greater distance from the opponent and uses projectiles at a slightly higher rate compared to the default reward agent. This demonstrates that our modularized reward function, incorporating customized reward terms, effectively guides DRL agents to learn the intended distinct play style.}
    \label{fig:result_defensive_type}
\end{figure}
Adjusting the reward terms listed in TABLE~\ref{table_reward_terms_default}, we obtain agents who exhibit various game styles. As an example, Fig.~\ref{fig:result_defensive_type} compares a defensive agent, trained using reward terms settings that favor defensive play, with an agent trained with the default reward function. Two key behavioral differences are observed: in Fig.~\ref{fig:result_defensive_type}~(a)the number of projectiles used per match, and Fig.~\ref{fig:result_defensive_type}~(b) the average distance from the opponent, the defensive agent demonstrates a tendency to maintain a longer distance from the opponent and rely on projectile attacks slightly more. This confirms that our model and reward function effectively guide the learning of specialized combat styles.

\subsubsection{Diverse Opponents}
By modifying the reward parameters, we successfully train a variety of DRL agents with different game-playing styles. In addition to the special move and defensive types discussed above, we develop agents with air-based (favoring aerial attacks), cowardly (reluctant to engage in combat), and other distinct behaviors. Furthermore, to accommodate novice players, we introduce agents with newbie and key-spamming behaviors. 

\subsubsection{Evaluation of LLM Output Quality}

\begin{figure}
    \centering
    \includegraphics[width=1.0\linewidth]{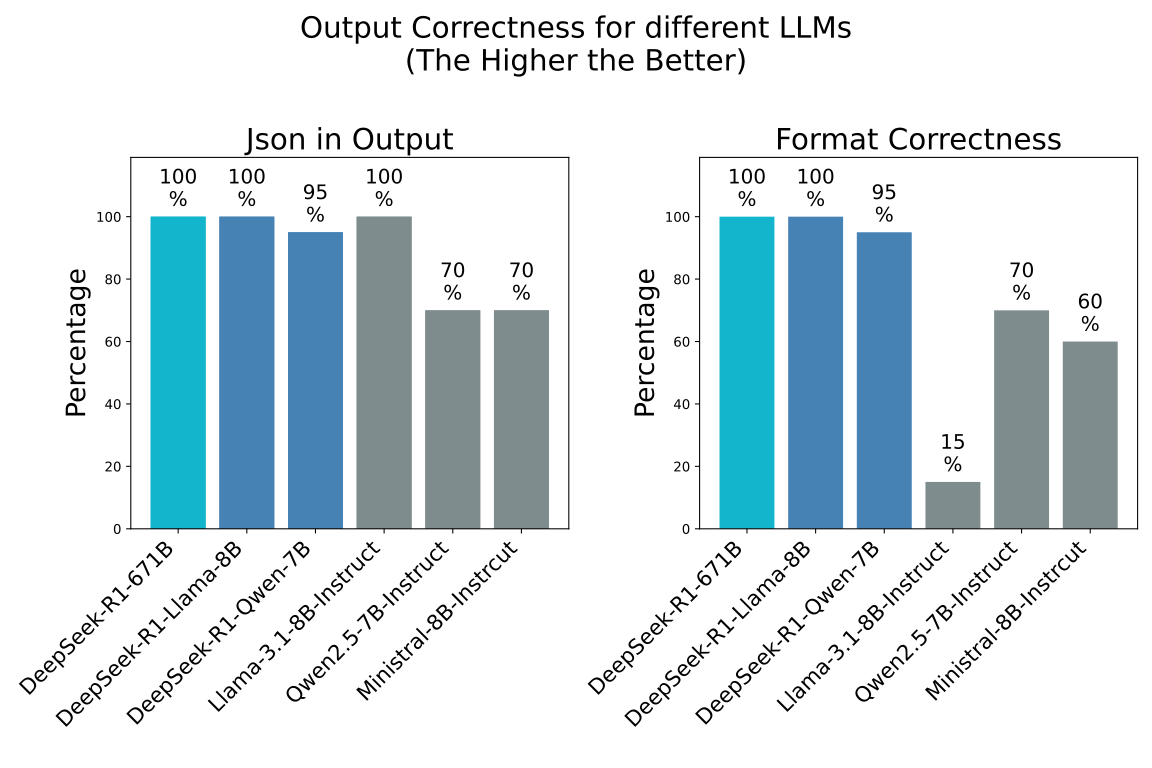}
    \caption{Correctness of LLM output across models. "JSON in Output" indicates whether a valid JSON was generated; "Format Correctness" further requires CoT reasoning and only one JSON in the output. Reasoning-optimized models significantly outperform non-reasoning models.}
    \label{fig:result_llms_correctness}
\end{figure}
\begin{figure}
    \centering
    \includegraphics[width=1.0\linewidth]{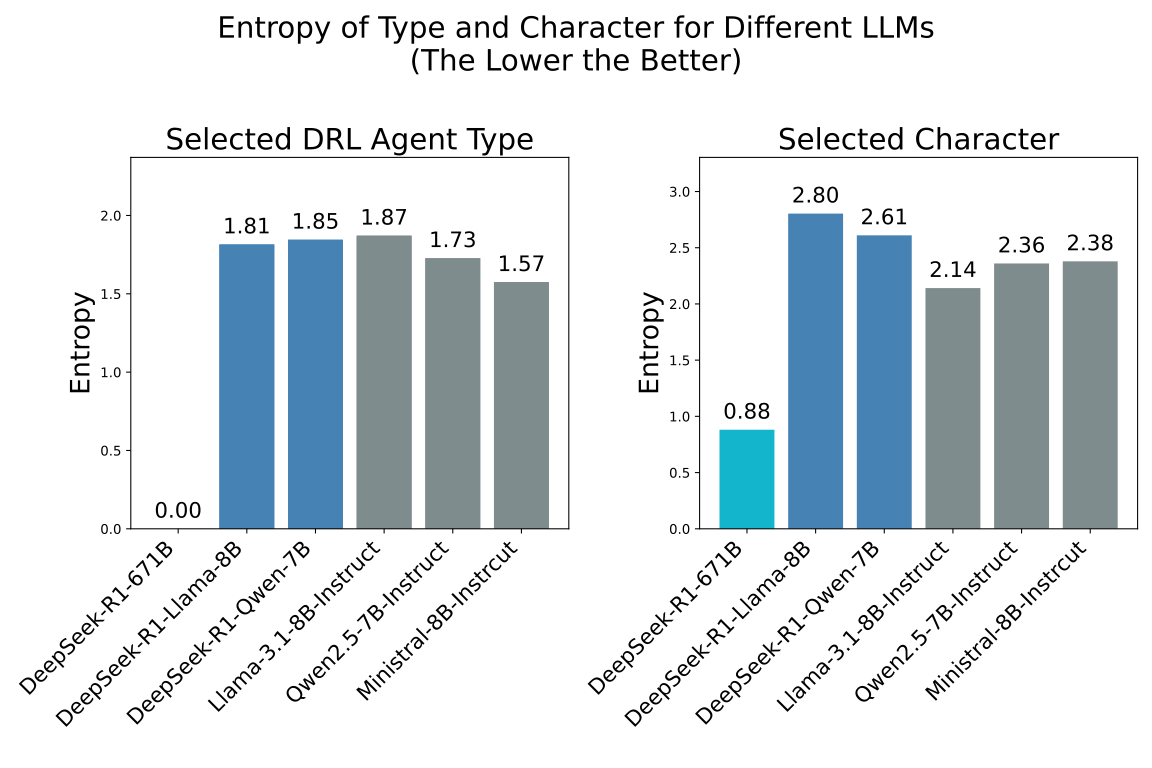}
    \caption{Entropy of selections for opponent's type and character across 20 responses. Lower entropy indicates more consistent and reliable opponent choices. Only valid outputs (containing JSON) were included in the computation. The full-scale DeepSeek-R1-671B model significantly outperforms all other models, and the differences in performance among the other models are marginal.}
    \label{fig:result_llms_entropy}
\end{figure}

To evaluate the performance of different LLMs within the LLMHA framework, we conducted 20 inference rounds using identical input prompts for each model. We designed two evaluation metrics—\textbf{Correctness} and \textbf{Entropy}—to assess the output quality.

\textbf{Correctness} is composed of two indicators: \textit{JSON in Output} and \textit{Format Correctness}. The former measures whether the LLM output contains a valid JSON block from which the selected opponent can be extracted; if missing, the response is considered a failure. The latter checks whether the output conforms to two key format requirements: the inclusion of both only one JSON block and CoT (Chain-of-Thought) reasoning. Although outputs that violate this format may still yield an extractable opponent, they are deemed to be lower quality.

\textbf{Entropy} specifically Shannon entropy, measures the diversity of LLM selections over 20 generations and includes two components: entropy over \textit{ DRL Agent Types Selected} and entropy over \textit{Characters Selected}. We assume that a lower entropy indicates a more consistent and stable selection behavior, which reflects better reasoning quality. In contrast, a higher entropy implies greater randomness and lower reliability. Note that outputs that fail the JSON requirement are excluded from the entropy calculation.

As shown in Fig.~\ref{fig:result_llms_correctness}, reasoning-optimized models (e.g., DeepSeek-R1 variants) significantly outperform non-reasoning models in terms of correctness. Fig.~\ref{fig:result_llms_entropy} shows that the entropy values are generally similar across most models except the full-scale DeepSeek-R1-67B, which exhibits notably more consistent selections. Although smaller models do not show strong differences in entropy, the higher correctness scores of reasoning models suggest that they are more reliable overall.

In conclusion, full-scale DeepSeek-R1-671B is the best performing model for LLMHA. For cost-sensitive deployment, DeepSeek-R1-Llama-8B offers slightly better correctness than DeepSeek-R1-Qwen-7B with comparable entropy, making it a more favorable low-cost alternative. Non-reasoning models are not suitable for use as LLMHA due to their low output reliability.

\subsubsection{User Study}
\label{Exp: user_study}
\begin{figure}
    \centering
    \includegraphics[width=1.0\linewidth]{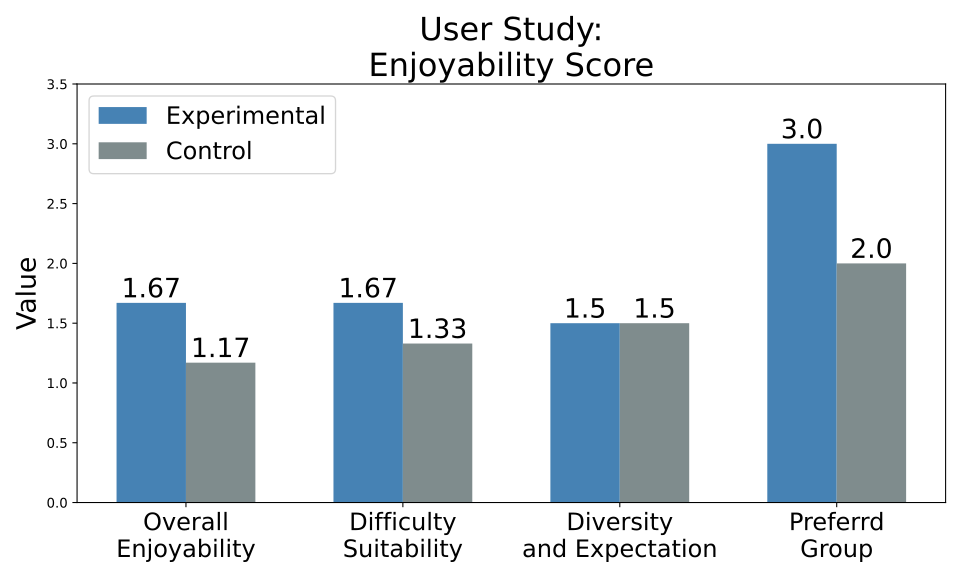}
    \caption{Results of the user study on subjective enjoyment ratings. The figure presents the mean scores for Overall Enjoyability, Difficulty Suitability, Diversity and Expectation, and Preferred Group, where higher values indicate greater player satisfaction. The experimental group used the TTA system with LLMHA-selected opponents, while the control group had opponents assigned randomly. The results show that the TTA system significantly enhances player enjoyment compared to random selection. The only metric where both groups achieved similar scores was Diversity and Expectation, likely because the control group also sampled from the same agent archive which has diverse DRL agents.}
    \label{fig:result_enjoyability_score}
\end{figure}

Although quantitative analysis has demonstrated the effectiveness of our method, the enjoyment of the player ultimately manifests itself as a subjective experience. Therefore, we conducted a small-scale user study involving human players to evaluate the impact of our approach.

\textbf{Study Design:}  
We divided the opponents into a control group and an experimental group, each group consisting of five matches. In the experimental group, LLMHA selected deep reinforcement learning (DRL) agents from the agent archive as the opponent, while in the control group, the game master (GM) randomly selected agents from the archive.

Before the experiment, participants received a questionnaire that included two types of questions: some to be answered immediately after each match and others to be answered after completing a full set of matches in either the control or experimental group. All questions were designed as single-choice questions. During the experiment: Since LLMHA required inference time to select the next opponent in the experimental group, before each match began, a random waiting period of approximately 1 to 2 minutes was introduced in the control group to simulate this delay. After the match, players were invited to provide feedback in natural language, which was processed by the GM. After submitting their questionnaire responses for that round, they proceeded to the next match. Upon completing all five matches in each group, they answered an additional set of questions that evaluated their overall experience. 
    
\textbf{Evaluation Metrics:}  
After the experiment, we computed a set of scores that measure the enjoyment of each group based on the responses of the players. For example, in the question ``How pleasant was the overall experience of this match group?" responses were assigned numerical scores: very enjoyable = 3, somewhat enjoyable = 2, neutral = 1, not enjoyable = 0. Similar scoring was applied to other questions that assessed other aspects of the enjoyment of the players.

Our questionnaire was designed to evaluate key factors, such as player enjoyment and difficulty of the match. Although independently developed, it aligns with established frameworks like the Player Experience of Need Satisfaction (PENS)\cite{PENS_johnson2018validation}, particularly in assessing ``competence" (challenge-skill balance) and ``autonomy".

\textbf{Results and Analysis:}  
The participants included casual gamers and some people with minimal gaming experience. Due to the small sample size with only six participants, we did not perform statistical significance tests but instead directly compared the mean values of a set of scores to measure the enjoyment of players. Future research can explore larger-scale studies with statistical testing for a more robust validation.

As shown in Fig.~\ref{fig:result_enjoyability_score}, the histograms illustrate the Enjoyability Scores for both the experimental and control groups, demonstrating that the TTA system with LLMHA led to notable improvements in the Overall Enjoyability, Difficulty Suitability, and Preferred Group ratings. The players in the experimental group reported greater enjoyment, indicating that the dynamically selected opponents improved their experience. The improved Difficulty Suitability suggests that LLMHA effectively matched players with appropriate opponents, preventing matches from being too easy or too difficult. Additionally, a higher Preferred Group score reflects that more players favored the TTA system over random selection, reinforcing the effectiveness of our approach in optimizing player satisfaction.

However, Diversity and Expectation did not show significant differences between the experimental and control groups. After thorough analysis, we conclude that this outcome is likely due to the fact that both groups sampled from the same diverse agent archive, which contains a wide variety of DRL agents with distinct play styles. Although the control group randomly selected opponents, the inherent diversity of the agent archive ensured that players still encountered a wide range of opponents, leading to comparable diversity scores. This may explain why the results deviated from our initial expectations.

\section{Discussion}
\subsection{Performance Against Specific Strategies}
Our DRL agents demonstrate strong performance against many strategies, particularly in close-range (melee) combat. However, they struggle against opponents that rely heavily on long-range attacks, such as projectiles. We hypothesize that this limitation stems from the agent's inherent strengths and tactical preferences. Specifically, agents benefit from (1) fast reaction times and input speed, allowing them to dominate melee engagements, and (2) frequent use of special moves, which typically inflict high damage and offer brief invincibility frames, granting a significant advantage in close-quarters combat.

However, overreliance on special moves introduces vulnerability. After the invincibility period ends, the agent enters a recovery state during which the character cannot act or defend—commonly referred to as "recovery frames" in fighting games. These recovery frames usually become a liability when faced with ranged opponents who can exploit this vulnerability with projectiles. Thus, the agent's close-combat specialization and reliance on special moves hinder its adaptability to distance-based and some other strategies.
\subsection{Effectiveness of Advanced Skill Usage}
While our model successfully learns to use special moves, the execution lacks precision. We found that, the number of special moves increases during training, but the number of regular attacks also rises, though to a lesser extent. Ideally, if the agent truly masters special moves, the number of regular attacks should remain relatively stable while the number of special moves increases, and the two should eventually be of similar magnitude. However, our results show both increasing together.

We hypothesize that the agent learns an approximate action pattern for special moves rather than mastering the precise input sequences and timing windows. For example, to trigger a Hadouken, the agent may repeatedly input combinations of "down," "forward," and "punch" in hopes of activating the move, rather than executing the exact command reliably.

To address this limitation, future work could improve the model architecture for extracting temporal information. Enhancing the current LSTM design or introducing transformer-based attention mechanisms\cite{attn_is_all_you_need_vaswani2017attention} may enable the agent to better capture and utilize precise temporal dependencies.

\subsection{Limitations in Advanced Playing Strategies}

In high-level human matches, players often employ advanced tactical strategies that go beyond individual skills, such as special moves. Some examples of these strategies are as follows.

\begin{itemize}
    \item{Spacing:} The players' deliberate control of space and set a situation that would be advantageous to them, typically involving movement and neutral attacks to keep the opponent in check.
    \item{Link Combos:} Sequences in which one move connects to another without canceling the previous action, typically by exploiting frame advantages between attacks.
    \item{Hit Confirm:} A technique in which players start a move sequence and only complete it after confirming a successful hit, allowing safe transitions into combos while avoiding punishment if blocked.
\end{itemize}

These strategies reflect not only a deep understanding of character mechanics, but also anticipate the opponent's reactions, grounded in human psychology and physiological limits. However, our DRL agents have not yet demonstrated the ability to utilize such high-level strategies.

We hypothesize three potential reasons for this limitation:
\begin{enumerate}
    \item \textbf{Insufficient training time.} Although no trend has been observed, extended training with increased opponent diversity may eventually lead to emergent behaviors, as demonstrated in prior work on emergent bartering strategies in multi-agent reinforcement learning\cite{johanson2022emergent}

    \item \textbf{Strategic irrelevance for DRL agents.} Techniques like spacing and link combos are often designed to exploit limitations in human cognition and reaction speed. For DRL agents that do not share these constraints, such tactics may offer limited utility.

    \item \textbf{Architectural limitations.} While our network design supports learning advanced skills such as special moves, it may still lack the capacity to capture more complex strategies like spacing and link combos. Drawing a parallel to emergent capabilities observed in large language models\cite{wei2022emergent}, we speculate that similar emergent behaviors may arise in DRL agents given sufficiently well-designed architectures and representational power.
\end{enumerate}
Future work will explore extended training durations, opponent diversity, and improved model architectures to investigate whether such advanced strategies can emerge under the right conditions.

\subsection{Action Space}
In our experimental setup, the agent's action is represented as a 12-dimensional multi-binary array, where each element corresponds to a specific arcade controller button. We suspect that this action space may be overly simplistic for the complexities of Street Fighter II. As mentioned earlier, while our agents are capable of performing advanced techniques such as special moves, their understanding and appropriate application of these moves remain limited.

One potential solution is to adopt Hierarchical Reinforcement Learning (HRL), where the agent first learns the low-level action sequences required to execute special moves and then learns when and how to use them. However, modeling such behavior in Street Fighter II is highly challenging due to the nuanced nature of its strategies. For example, charge-based characters like M.Bison often hold back and crouch for extended periods as a preparatory step to charge multiple special moves. The actual move executed depends heavily on the opponent's actions and the evolving dynamics of the match, making it difficult to define clear high-level options for HRL.

Another possible direction is to learn more expressive action representations. This may enhance the agent's understanding and execution of advanced techniques. Prior research has shown that learning action representations can improve generalization in environments with large and complex action spaces~\cite{action_repre_chandak2019learning}.
\subsection{LLMHA's Open-Ended Output}
Autoregressive LLMs do not inherently guarantee the generation of legal outputs for opponent selection. For instance, the model may occasionally fail to output a valid dictionary structure. Although our LLMHA implements an output validation mechanism that prompts the model to regenerate responses until a valid output is produced, this retry process can introduce noticeable delays, potentially reducing enjoyment.

We also observed that smaller models, such as DeepSeek-R1-Distill-Qwen-7B, often struggle with long-context understanding, especially compared to their larger counterparts. Specifically, when the prompt contains many few-shot learning examples with example playing data, the model may mistakenly treat the examples as real player data, leading to incorrect opponent selections.

To address these issues, more carefully engineered prompt templates are needed. Furthermore, instruction tuning could improve the model’s ability to serve as a Hyper Agent. For potential deployment in commercial games, player feedback mechanisms, such as rating the LLMHA's selectionsdecisions, could be incorporated into the game environment. These ratings would facilitate reinforcement learning from human feedback (RLHF)\cite{rlhf_ouyang2022training} to further fine-tune the LLM as a Hyper Agent.

\section{Conclusion}
In this paper, we proposed a two-tier agent (TTA) system to enhance player enjoyment in fighting games, focusing on SF2 as our evaluation environment. The first tier employed a specifically designed network architecture, modularized reward functions, and hybrid training methods to create diverse DRL agents with distinct playing styles. In the second tier, a Large Language Model Hyper-Agent dynamically selected suitable opponents based on players' gameplay data and feedback. Experimental results demonstrated significant improvements in agent capabilities in performing advanced skills, special move (156. 36\%) and player enjoyment (42.73\%), confirming the TTA system's effectiveness.

\section{Future Works}
Future work will explore several promising directions. First, we plan to enhance the adaptability of our DRL agents to long-range combat scenarios by explicitly modeling recovery frames and opponent projectile patterns. Second, improving the temporal precision of advanced skill execution through architectures such as Transformer-based models will be investigated. In addition, extended training durations and increased opponent diversity will be used to encourage the emergence of high-level strategic behaviors, including spacing and hit-confirm tactics. Finally, refining the LLM Hyper-Agent via instruction tuning, reinforcement learning from human feedback (RLHF), and improved prompt engineering will further enhance system responsiveness and enjoyment in practical deployments.

\bibliography{references}   
\bibliographystyle{IEEEtran}
\appendices

\section{Training Parameters}

\begin{table}[H]
    \caption{Training parameter}
    \begin{center}
        \begin{tabular}{c c|c c}
            \hline
            \multicolumn{2}{c|}{\textbf{PPO}} & \multicolumn{2}{|c}{\textbf{Hybrid training}}\\
            \hline 
            \textbf{Parameters} & \textbf{Values} & \textbf{Parameters} & \textbf{Values} \\
            \cline{1-4}
            {vf coef} & {1.0} & {Steps per iteration} &  {$5 \times 10^{6}$}\\
            {ent coef} & {0.01} & {Self-play ratio} & {0.7} \\
            {n\_steps} & {512} & {num\_envs} & {12} \\
            {batch\_size} & {256}  & {Character flip rate} & {0.5} \\
            {PPO $\gamma$} & {0.1} & {Policy pool update} & {All} \\
            {lr initial value} & {$2.5 \times 10^{-4}$} &  {Opponent Selection} & {All} \\
            {lr final value} & {$2.5 \times 10^{-6}$}  &  {} & {} \\
            {Optimizer} & {Adam} &  {} & {} \\
            \hline
        \end{tabular}
        \begin{flushleft}
            \footnotesize
            Some commonly known parameters (e.g., batch size, optimizer) are omitted from detailed descriptions. 
            $^{\mathrm{a}}$ vf\_coef: Value function coefficient for loss calculation.  
            $^{\mathrm{b}}$ ent\_coef: Entropy coefficient for loss calculation.  
            $^{\mathrm{c}}$ lr initial value / lr final value: We use a linear learning rate scheduler where the learning rate decreases linearly from the initial to the final value during training.  
            $^{\mathrm{d}}$ Self-play ratio: The proportion of self-play tasks among all training tasks; the remaining tasks are PvE tasks.  
            $^{\mathrm{e}}$ num\_envs: Number of parallel environments used during training. Increasing this value speeds up training but requires more computational resources.  
            $^{\mathrm{f}}$ Character Flip Rate: Probability that the DRL agent and the opponent swap roles in self-play tasks (e.g., Player 1 $\leftrightarrow$ Player 2).  
            $^{\mathrm{g}}$ Policy Pool Update: Determines how the policy pool is updated. ``top\_N'' retains only the strongest $N$ models, while ``All'' retains all models.  
            $^{\mathrm{h}}$ Opponent Selection: Defines how self-play opponents are chosen. ``top\_N'' selects the strongest $N$ models as opponents, while ``All'' selects all models in the policy pool.  
        \end{flushleft}
        \label{training_hyperparameter}
    \end{center}
\end{table}

\section{Reward Terms for DRL Agents} \label{appd_reward_term_settings}
TABLE~\ref{table_reward_terms} provides the reward term settings for 7 types of DRL agents.
\begin{table*}[htbp]
            \caption{Reward Terms}
            \begin{center}
                \begin{tabular}{c|c|c|c|c|c|c|c}
                    \hline 
                    \textbf{Parameters} & \textbf{Default} & \textbf{Special Move} & \textbf{Defensive} & \textbf{Air}& \textbf{Newbie} & \textbf{Coward} & \textbf{Key Spamming}\\
                    \hline
                    {Reward scale} & {0.001} & {0.001} & {0.001} & {0.001} & {0.001}  & {0.001} & {0.001}\\
                    {Raw reward coefficient} & {1.0} & {1.0} & {1.0} & {1.0} & {1.0} & {1.0}  & {0.0} \\
                    {Special move bonus} & {1.0} & {3.0} & {1.0}  & {0.0} & {0.0} & {0.0} & {0.0}\\
                    {Projectile bonus} & {1.0} & {1.0} & {1.0} & {0.0} & {0.0} & {0.0} & {0.0} \\
                    {Distance bonus} & {1.0} & {2.0}&  {0.0} & {0.0} & {0.0} & {0.0} & {0.0} \\
                    {Special move reward} & {0.0} & {10.0}& {0.0} & {0.0} & {0.0} & {0.0} & {0.0} \\
                    {Projectile reward} & {0.0} & {0.0}& {10.0}  & {0.0} & {0.0} & {0.0} & {0.0}\\
                    {Distance reward} & {0.0} & {0.0} & {0.02} & {0.0} & {0.0} & {0.2} & {0.0} \\
                    {In air reward} & {0} & {0.0}& {0.0}  & {0.05} & {0.0} & {0.0} & {0.0}\\
                    {Time reward} & {0.0} & {0.0} & {0.0} & {0.0} & {0.0} & {0.0} & {0.0} \\
                    {Cost coefficient} & {0.0} & {1.0} & {0.0} & {0.0} & {1.0} & {3.0} & {-1.0} \\
                    {Special move cost} & {0.0} & {0.0} & {0.0} & {0.0} & {30.0} & {5.0} & {-3.0}\\
                    {Regular attack cost} & {0.0} & {1.0} & {0.0} & {0.0} & {0.0} & {1.0} & {2.0}\\
                    {Jump cost} & {0.0} & {0.0} & {0.0} & {0.0} & {0.0} & {0.0} & {0.0}\\
                    {Vulnerable frame cost} & {0.0} & {0.05}  & {0.0} & {0.0} & {0.0} & {0.0} & {0.0}\\
                    \hline
                \end{tabular}
            \label{table_reward_terms}
            \end{center}
\end{table*}

\section{Prompt for LLMHA} \label{appd_prompt_templates}

The prompt template serves as the foundation of the prompt construction and is composed of multiple modular sections: SELECTION PRINCIPLES, OUTPUT FORMAT REQUIREMENT, PLAYING DATA, ARCHIVE INFO, and FEW SHOT EXAMPLES. Each section is populated with the corresponding content and then concatenated into a single prompt, which is fed into the LLMHA as input.

For the FEW SHOT EXAMPLES module, one or more examples can be optionally included. However, as previously noted, smaller-scale models may mistakenly treat these examples as actual playing data. To mitigate this issue and enhance compatibility with smaller models (e.g., DeepSeek-R1-Qwen-7B), we also design a Simplified In-Context Learning Example variant that excludes all data from the example.

\begin{tcolorbox}[breakable, colback=gray!5!white, colframe=gray!75!black, title=Prompt Template]
You are a "Hyper Game-Playing Agent" large language model. Your goal is to choose the next opponent, which is a trained deep reinforecement learning (DRL) model, for a human player in video game, Street Fighter II, in order to enhance their overall enjoyment. You must make a decision that balances difficulty, the player's recent performance, diversity of opponents, and other factors described below.

1) First, you should follow the Principles for Opponent Selection:\\
\{SELECTION\_PRINCIPLES\}\\

2) Second, you should follow the Output Format Requirements:\\
\{OUTPUT\_FORMAT\_REQUIREMENT\}\\

3) Third, you should understand and utilize the Playing Data Overview:\\
\{PLAYING\_DATA\}\\

4) Fourth, you should select an opponents from the Agent Archive.\\
Be aware, it is the model decides the difficulty but not suggested character for that model. The suggested characters means those characters are suitable for the corresponding type.
The dictionary below is the Agent Archive (It is stored in a dictionary format):\\
\{ARCHIVE\_INFO\}\\

5) Here is an Q\&A example for your reference:\\
\{FEW\_SHOT\_EXAMPLES\}\\

(Explicitly instruct the model to respond in valid JSON format, containing mandatory fields:
  - "chosen\_agent\_type"\\
  - "chosen\_agent\_model\_path"\\
  - "chosen\_agent\_character"\\
)\\
Important Note:
1. Output only the Chain of Thought reasoning and one JSON object in your answer.Do not include any additional contents.\\
2. Include the three required fields: "chosen\_agent\_type", "chosen\_agent\_model\_path", and "chosen\_character".\\
3. You should include Chain of Thought reasoning parts, but they mush be OUTSIDE of and BEFORE the JSON object. Make sure your Chain of Thought Reasoning is concise. THE LENGTH OF YOUR TOTAL OUTPUT SHOULD NOT BE MORE THEN 300 WORDS!!!\\

Now, based on the data and guidelines above, please provide your final decision in strict format:
\end{tcolorbox}

\begin{tcolorbox}[breakable, colback=gray!5!white, colframe=gray!75!black, title=Selection Principles]

a). Difficulty Adjustment (the most important):

   - If the player's win rate is very low or they are on a losing streak, choose an easier opponent.
   
   - If the player’s win rate is high or they are on a winning streak, choose a stronger opponent.
   
   - Overall, try to make the player wins, but still challenging.\\

b). Opponent Diversity:

   - Prefer an agent type and character that the player has not yet faced or has faced only a few times.
   
   - Rotate among different play styles (defensive, projectile-heavy, rushdown, etc.) so the player experiences variety.\\

c. Player Behaviors:

   - If the player rarely uses special moves (e.g., < 2 per match), assume they are a beginner and avoid overwhelming them with advanced AI.
   
   - If the player often uses combos or advanced techniques, they are more skilled; pick a challenging agent to maintain fun.\\

d). Agent/Character Features:

   - When possible, select agents that demonstrate distinct strategies (e.g., projectile spamming, grappling, or strong aerial attacks).
   
   - Avoid repeatedly using the same type/character unless necessary for balancing difficulty.\\

e). \MakeUppercase{consider player's feedback: }

    \MakeUppercase{last but not least, it is the most important} Check ``player's\_feedback'' and ``the\_last\_opponents'' in the provided ``playing\_data'' dictionary. If they provided any feedback, consider how they feel and what they suggest.
\end{tcolorbox}
\begin{tcolorbox}[breakable, colback=gray!5!white, colframe=gray!75!black, title=Output Format Requirement, breakable]
Please respond in strict JSON format with the following mandatory fields:
\\
- "chosen\_agent\_type" (string)\\
- "chosen\_agent\_model\_path" (string)\\
- "chosen\_agent\_character" (string)\\

All the three above must be present inside of the JSON object. You should include Chain of Thought reasoning parts, but they mush be OUTSIDE of and BEFORE the JSON object. Make sure your Chain of Thought Reasoning is concise. THE LENGTH OF YOUR TOTAL OUTPUT SHOULD NOT BE MORE THEN 300 WORDS!!!\\
Only output the Chain of Thought reasoning and one JSON object in your final answer, do not include any additional contents.

\end{tcolorbox}
\begin{tcolorbox}[breakable, colback=gray!5!white, colframe=gray!75!black, title=An Example of Playing Data]
\{\\
  "current\_character": "Ryu",\\
  "total\_matches": 6,\\
  "win\_rate": 0.8333333333333334,\\
  "total\_wins": 5,\\
  "total\_losses": 1,\\
  "current\_win\_streak": 3,\\
  "current\_loss\_streak": 0,\\
  "average\_score\_per\_match": "63/100",\\
  "average\_special\_moves\_per\_match": 9.5,\\
  "faced\_agents\_times": \{\\
    "projectile\_type": 3,\\
    "special\_move\_type": 0,\\
    "defensive\_type": 1,\\
    "aggressive\_type": 1,\\
    "air\_type": 1,\\
    "coward\_type": 0,\\
    "newbie\_type": 0,\\
    "key\_spamming\_type": 0\},\\
  "faced\_characters\_times": \{\\
    "Ryu": 2,\\
    "Ken": 0,\\
    "Chunli": 0,\\
    "Guile": 0,\\
    "Blanka": 0,\\
    "Zangief": 0,\\
    "Dhalsim": 0,\\
    "Balrog": 0,\\
    "Vega": 2,\\
    "Sagat": 1,\\
    "Bison": 0,\\
    "EHonda": 1\},\\
  "the\_last\_opponents": \{\\
    "type": "aggressive\_type",\\
    "character": "EHonda",\\
    "model\_path": \\
    "agent\_models/agents\_archive/aggressive \_type\allowbreak /1\_0.22",\\
    "difficulty": "8/10-(Hard)"\\
  \},\\
  "player's\_feedback": "This match is too simple, the enemy didn't perform any effective attack at all"\\
\}
\end{tcolorbox}

\begin{tcolorbox}[breakable, colback=gray!5!white, colframe=gray!75!black, title=An Example of Archive Info]
\{\\
\ \ \ \ "projectile\_type": \{\\
\ \ \ \ \ \ \ \ "suggested\_characters\_for\_this\_type": [\\
\ \ \ \ \ \ \ \ \ \ \ \ "Ryu",\\
\ \ \ \ \ \ \ \ \ \ \ \ "Ken",\\
\ \ \ \ \ \ \ \ \ \ \ \ "Sagat",\\
\ \ \ \ \ \ \ \ \ \ \ \ "Dhalsim"\\
\ \ \ \ \ \ \ \ ],\\
\ \ \ \ \ \ \ \ "agent\_models": [\\
\ \ \ \ \ \ \ \ \ \ \ \ \{\\
\ \ \ \ \ \ \ \ \ \ \ \ \ \ \ \ "model\_path": "agent\_models/agents\_archive\allowbreak /projectile\_type/2\_0.2",\\
\ \ \ \ \ \ \ \ \ \ \ \ \ \ \ \ "model\_difficulty\_score": "6/10-(Medium)"\\
\ \ \ \ \ \ \ \ \ \ \ \ \}\\
\ \ \ \ \ \ \ \ ]\\
\ \ \ \ \},\\
\ \ \ \ "special\_move\_type": \{\\
\ \ \ \ \ \ \ \ "suggested\_characters\_for\_this\_type": [\\
\ \ \ \ \ \ \ \ \ \ \ \ "Ryu",\\
\ \ \ \ \ \ \ \ \ \ \ \ "Ken",\\
\ \ \ \ \ \ \ \ \ \ \ \ "Sagat",\\
\ \ \ \ \ \ \ \ \ \ \ \ "Dhalsim"\\
\ \ \ \ \ \ \ \ ],\\
\ \ \ \ \ \ \ \ "agent\_models": [\\
\ \ \ \ \ \ \ \ \ \ \ \ \{\\
\ \ \ \ \ \ \ \ \ \ \ \ \ \ \ \ "model\_path": "agent\_models/agents\_archive\allowbreak /special\_move\_type/1\_0.25",\\
\ \ \ \ \ \ \ \ \ \ \ \ \ \ \ \ "model\_difficulty\_score": "9/10-(Hard)"\\
\ \ \ \ \ \ \ \ \ \ \ \ \}\\
\ \ \ \ \ \ \ \ ]\\
\ \ \ \ \},\\
\ \ \ \ "newbie\_type": \{\\
\ \ \ \ \ \ \ \ "suggested\_characters\_for\_this\_type": [\\
\ \ \ \ \ \ \ \ \ \ \ \ "Ryu",\\
\ \ \ \ \ \ \ \ \ \ \ \ "Ken",\\
\ \ \ \ \ \ \ \ \ \ \ \ "Dhalsim",\\
\ \ \ \ \ \ \ \ \ \ \ \ "EHonda",\\
\ \ \ \ \ \ \ \ \ \ \ \ "Chunli",\\
\ \ \ \ \ \ \ \ \ \ \ \ "Blanka",\\
\ \ \ \ \ \ \ \ \ \ \ \ "Guile",\\
\ \ \ \ \ \ \ \ \ \ \ \ "Zangief",\\
\ \ \ \ \ \ \ \ \ \ \ \ "Balrog",\\
\ \ \ \ \ \ \ \ \ \ \ \ "Vega",\\
\ \ \ \ \ \ \ \ \ \ \ \ "Sagat",\\
\ \ \ \ \ \ \ \ \ \ \ \ "Bison"\\
\ \ \ \ \ \ \ \ ],\\
\ \ \ \ \ \ \ \ "agent\_models": [\\
\ \ \ \ \ \ \ \ \ \ \ \ \{\\
\ \ \ \ \ \ \ \ \ \ \ \ \ \ \ \ "model\_path": "agent\_models/agents\_archive\allowbreak /newbie\_type/1\_0.017",\\
\ \ \ \ \ \ \ \ \ \ \ \ \ \ \ \ "model\_difficulty\_score": "0/10-(Very Easy)"\\
\ \ \ \ \ \ \ \ \ \ \ \ \}\\
\ \ \ \ \ \ \ \ ]\\
\ \ \ \ \}\\
\}\\

\end{tcolorbox}

\begin{tcolorbox}[breakable, colback=gray!5!white, colframe=gray!75!black, title=In-Context Learning Example]

(This is the begining of In-context learning texts)

The Question Asked by User:\\

You are a ``Hyper Game-Playing Agent'' large language model. Your goal is to choose the next opponent for human players to make the game-playing more enjoyable to human players.
(The playing data and archives are omitted in this example.)\\

The Ideal Answer for the Question:
\#\#\# Reasoning
(BE AWARE! THE DATA MENTIONED IN THE EXAMPLE IS NOT TRUE DATA, JUST AN EXAMPLE!!)
The player's overall win rate is 0.38 with an average of 1.5 special moves per match, indicating a beginner. To avoid overwhelming them, lower-score AIs are preferable. \\

They've already faced Ryu, Balrog, Ken, Sagat, Chunli, and EHonda, so we should consider new types/characters. ``Projectile\_type'' with a -0.17 score is weaker, and Sagat is an available character the player hasn't fought much. \\
Thus, ``lstm\_projectile2'' controlling Sagat is a suitable choice for balanced difficulty and novelty.\\

\#\#\# JSON\\
\{\\
    ``chosen\_agent\_type'': ``projectile\_type'',\\
    ``chosen\_agent\_model\_path'': ``agent\_models/agents\_archive/projectile\_type\allowbreak 
 /lstm\_projectile2'',\\
    ``chosen\_agent\_character'': ``Sagat''\\
\} \\
(This is the end of In-context learning texts)
\end{tcolorbox}

\begin{tcolorbox}[breakable, colback=gray!5!white, colframe=gray!75!black, title= Simplified In-Context Learning Example]
(This is the begining of In-context learning texts)

The Question Asked by User:\\

You are a "Hyper Game-Playing Agent" large language model. Your goal is to choose the next opponent for human players to make the game-playing more enjoyable to human players.
(The playing data and archives are omitted in this example.)\\

The Ideal Answer for the Question:
(Your reasoning ... ...)\\
\#\#\#json\\
\{\\
    "chosen\_agent\_type": "projectile\_type",\\
    "chosen\_agent\_model\_path": "agent\_models/agents\_archive/projectile\_typee\allowbreak /lstm\_projectile2",\\
    "chosen\_agent\_character": "Sagat"\\
\} 

(This is the end of In-context learning texts)

\end{tcolorbox}

\section{LLMHA Output Example} \label{appd_llmha_output}

\begin{tcolorbox}[breakable, colback=gray!5!white, colframe=gray!75!black, title= LLMHA Output Example]
chain of thought followed by the JSON.\\
\verb|</think>|\\

To determine the best opponent for the player, I analyze their performance and feedback. The player has a high win rate (83.33\%) and a current win streak of 3, indicating they're performing well. However, their feedback suggests the last match was too simple, implying they'd benefit from a more challenging opponent this time. \\

Considering their character, Ryu, and the types they've faced, the next opponent should offer a different playstyle. The aggressive type, particularly EHonda, fits well. EHonda is aggressive and can provide a good challenge without overwhelming the player, aligning with the goal of maintaining fun and difficulty balance.\\

The aggressive\_type agent model at ``agent\_models/agents\_archive/aggressive\_type/1\_0.22'' is chosen for its difficulty score of 8/10, ensuring a challenging yet manageable match for the player.

\#\#\# JSON\\
```json\\
\{\\
    ``chosen\_agent\_type'': ``aggressive\_type'',\\
    ``chosen\_agent\_model\_path'': ``agent\_models/agents\_archive/aggressive\_type\allowbreak /1\_0.22'',\\
    ``chosen\_agent\_character'': ``Honda''\\
\}\\
```
\end{tcolorbox}

\color{red}

\end{document}